\documentclass{article}
\usepackage{spconf,amsmath,epsfig}

\title{Local Background Estimation for Improved Gas Plume Identification in Hyperspectral Images}

\threeauthors
{Scout Jarman\sthanks{The authors acknowledge the Aerospace Corporation for collecting and providing the historical airborne LWIR data from the Los Angeles basin area. This work is partially funded by NNSA Contract No. 89233218CNA000001, and Los Alamos National Laboratory subcontract No. C3486. LA-UR-24-20093.}}
{Utah State University\\Mathematics \& Statistics Dept.\\ Logan, UT}
{Zigfried Hampel-Arias, Adra Carr}
{Los Alamos National Laboratory\\Intelligence and Space Research\\	Los Alamos, NM}
{Kevin R. Moon}
{Utah State University\\Mathematics \& Statistics Dept.\\ Logan, UT}
\begin{document}
%\ninept
%
\maketitle
\begin{abstract}
Deep learning identification models have shown promise for identifying gas plumes in Longwave IR hyperspectral images of urban scenes, particularly when a large library of gases are being considered.
Because many gases have similar spectral signatures, it is important to properly estimate the signal from a detected plume.
Typically, a scene's global mean spectrum and covariance matrix are estimated to whiten the plume's signal, which removes the background's signature from the gas signature.
However, urban scenes can have many different background materials that are spatially and spectrally heterogeneous.
This can lead to poor identification performance when the global background estimate is not representative of a given local background material.
We use image segmentation, along with an iterative background estimation algorithm, to create local estimates for the various background materials that reside underneath a gas plume.
Our method outperforms global background estimation on a set of simulated and real gas plumes.
This method shows promise in increasing deep learning identification confidence, while being simple and easy to tune when considering diverse plumes.
\end{abstract}


\section{Introduction}
\label{sec:intro}

Hyperspectral imaging (HSI) has become increasingly popular due to its application in fields ranging from national security to medical research.
Longwave IR (LWIR) HSI has been shown to be particularly useful for the task of gas plume detection and identification \cite{Khan2018}.
This process involves first using a model to detect potential gas plumes followed by using a model to identify the gas in the plume.

There are several well established methods for identifying gases, such as Step-wise Linear Regression and Bayesian Model Averaging, which work well for identifying gases from a relatively small library of candidate gases \cite{Manolakis2014LongWaveIH}.
Machine learning and deep learning classification models have proven to be superior for identification, particularly when there are many candidate gases and many images to analyze \cite{Li2019, Klein2023HyperspectralTI, gewali2019machine}.
However, applying deep learning classifiers to real world data can be difficult, thus it is important to properly preprocess the data before classification.

Whitening is a common preprocessing step when working with nadir observed gas plumes.
Whitening uses the equation $\mathbf{z} = C^{-1/2}(\mathbf{x}-\mathbf{\mu})$, where $C$ is the scene covariance matrix, $\mu$ is the scene average spectral signature, and $\mathbf{x}$ is the spectrum of interest.
Typically a ``global" estimate of $\mu$ and $C$ are used, where every pixel in the scene, except for pixels in the region of interest (ROI), are included in the calculation.
However, urban images often contain many diverse background materials which are not represented well by the global $\mu$.
This can lead to mischaracterization of important identifying wavelengths for the gas, making it more difficult to classify correctly.

Since identification with deep learning models is already a difficult task on real world plumes, it is crucial to have an accurate estimate of the whitened gas signature.
This paper proposes an iterative background estimation approach with image segmentation to create local estimates of background spectra for plumes.
We demonstrate our approach to be beneficial in estimating gas signatures in both simulated and real world plumes when compared to the global whitening approach.

\section{Method}
\label{sec:method}

A standard approach for whitening is to assume that all spectral signatures in the image are normally distributed.
The normality assumption is not guaranteed to be satisfied, and in images with many different background materials, such as in urban scenes, the global background may not sufficiently represent the variety of background signatures \cite{Matteoli2014}.
Whitening with an incorrect background signature can lead to inaccurate whitened gas spectra, making it more difficult to classify the plume correctly.
For this reason, it is beneficial to whiten each part of a plume according to the distinct background materials it spans within the scene.

One option to distinguish between various background materials is to use image segmentation to break the image into groups of similar pixels.
We use Watershed Segmentation (WS) as it has an appropriate adaption for use with HSI \cite{Tarabalka2010SegmentationAC}.
WS determines segments by finding boundaries between ``different colored" regions, thus it is well suited for the purpose of distinguishing between various background materials.
We choose to use non-marker based WS which results in over-segmentation, reducing the chance that a given segment will contain more than one distinct background material.
Since some segments may have a relatively small number of pixels, we focus on locally estimating the means while using a globally estimated covariance matrix \cite{Jarman2023EnsembleSF}.

WS produces segments that are treated independently which can cause problems for segments that are completely masked by a plume, as whitening with the segment's mean will remove important characteristics belonging to the gas.
For this reason, we need to be able to estimate the background of a contaminated segment using non-contaminated pixels.
Inspired by \cite{Theiler2017LocalBE}, we represent each spectral signature $y$ with an additive model, $y=b+\alpha t$, where $b$ is the background signature, $t$ is the target gas signature, and $\alpha$ is the non-negative signal strength.

Given a set of pixels of the same background material, we want to estimate $b$ and $t$ which should be the same for all pixels in the set, and estimate $\alpha$ for each individual pixel.
One way to do this is by trying to solve the following optimization problem:
\begin{equation}
    \label{eq:min}
    \min_{b, t, \alpha_i\geq0} \frac{1}{N}\sum_i||y_i - (b+\alpha_it)||^2.
\end{equation}
This optimization problem minimizes the mean squared error (MSE) between each pixel and its modeled signature.
An issue with trying to solve this optimization problem is that $b$, $t$, and $\alpha_i$ are all unknown.
Thus, we opt to use an alternating approach to simplify the problem.
Using Lagrange multipliers, the solution at each step for $\alpha_i$, $b$, and $t$ are:
\begin{align}
    \alpha_i &= \max\left(0, \frac{(y_i - b)^T t}{t^Tt}\right),\label{eq:min_alpha}\\
    b &= \frac{1}{N} \sum y_i - \alpha_i t,\label{eq:min_b}\\
    t &= \frac{\sum (y_i-b)\alpha_i}{\sum \alpha_i^2}.\label{eq:min_t}
\end{align}

We call this algorithm iBATE (iterative Background Alpha Target Estimation).
Natural initializations for $b$ and $t$ are the global estimates, and $\alpha_i=1$.
Each step can be repeated until the error reaches a limit, or ceases to decrease; in nearly all of our test cases we observed convergence within 10 iterations.
Although iteratively solving these equations to minimize the error will produce an ``optimal" estimate for $b$, it will not produce a good estimate if all of the pixels are contaminated by the gas.
For this reason, and the observation that most segments are small or are completely masked by the gas, we need to incorporate additional pixels from other segments that are of the same background material but that are NOT contaminated by the gas.
Once such pixels are found, the initializations and optimizations are simplified since we can fix $\alpha_i=0$ for all non-plume pixels, and we can average the non-plume pixels to create a clean initial estimate of $b$.

To quantify segment similarity, we use linkage functions from Hierarchical Clustering, which are used to find the ``distance" between two collections of objects.
We propose the Truncated Average Linkage function (TAL), a modification of the Average Linkage function, where a proportion of the largest distances are removed \cite{Nielsen2016}.
TAL is defined as: 
\begin{equation}
    \label{eq:tal}
    TAL(A,B, \beta)=\frac{1}{N_\beta}\sum_{i=0}^{N_\beta}d(\Vec{a}, \Vec{b})_{(i)},
\end{equation} where $A$ and $B$ are two segments, $N_\beta=(1-\beta)|A||B|$, and $d(\Vec{a},\Vec{b})_{(i)}$ is the $i^\text{th}$ smallest distance between two pixels $\Vec{a}\in A$ and $\Vec{b}\in B$.
The primary benefit of TAL is that it is robust to comparisons between segments that have more than one distinct background material.
Additionally, as $\beta\to0$, TAL becomes Average Linkage, and as $\beta\to1$, TAL becomes Single Linkage, thus allowing us to compare the performance of three linkage functions at once.

TAL requires a distance function to compare two spectral signatures, of which there are many options \cite{Deborah2015ACE}.
However, pixels in a plume segment are contaminated with the gas, which will influence the distance calculated with a non-contaminated pixel.
We propose a distance function designed to be robust to contaminated signatures called Truncated Euclidean Distance (TED), which removes a proportion of wavelengths with the largest squared error.
TED is defined as
\begin{equation}
    \label{eq:ted}
    TED(\Vec{a}, \Vec{b}, \gamma) = \sqrt{\sum_{i=0}^{n_\gamma} (a-b)^2_{(i)}},
\end{equation} where $n_\gamma=(1-\gamma)c$, $c$ is the number of channels, and $(a-b)^2_{(i)}$ is the channel with the $i^\text{th}$ smallest squared difference.
Because gas signatures have relatively few wavelengths that are distinguishing, $\gamma$ can be chosen to account for these wavelengths, allowing proper comparison between contaminated and non-contaminated signatures.

We call the combination of the above methods LEBEAUS (Local itErative Background Estimation, Additive, Using Segmentation), which provides the ability to estimate various backgrounds for gas plumes in rural scenes.
Though LEBEAUS can be applied to any shape or size of plume, it is designed to be most useful for plumes that are subtended by either several background materials, or background materials that are not well represented by the global background estimate.

\section{Data \& Results}
\label{sec:data}

Because it is difficult to collect HSI with ground truth, it is useful to be able to simulate plumes where the ground truth is known.
To implant a gas signal into an individual pixel in a scene, we use the following equation:
\begin{equation}
    \label{eq:radtrans}
    L =  L_{bg} + \tau_a\left[B(T_p), -\epsilon_{bg}B(T_{bg})-\rho_{bg}L_d\right].
\end{equation}
For each pixel, $T_{bg}, L_d$, and $\tau_a$ are given, while $\epsilon_{bg}, T_{bg}$, and $\rho_{bg}$ are estimated.
A Gaussian plume dispersion model \cite{HOLMES2006} is used to simulate the concentration of the simulated plume, which is then used to define the shape and temperature of the plume.
For temperature, the relative concentration is used to smoothly transition from max stack temperature to ambient air temperature.

\begin{figure}[htb]
    \centering
    \includegraphics[width=.5\textwidth]{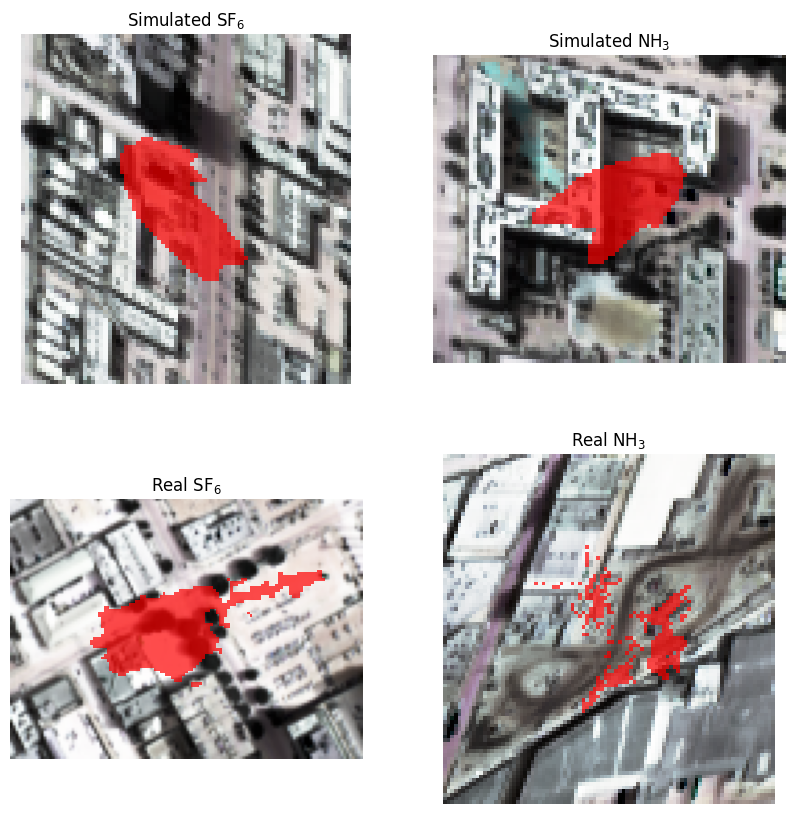}
    \caption{Plume ROIs shown in red over false-colored context images. First row are the simulated plumes, second row are the real plumes.}
    \label{fig:plumes}
\end{figure}

We simulated two large and dense plumes, one consisting of Sulfur Hexafluoride (SF$_6$) and another of Ammonia (NH$_3$).
SF$_6$ will act as a baseline plume as it is generally easier to identify compared to NH$_3$.
These plumes where implanted into the scene over background pixels that were found to be notably different from the global background signature.
In addition to these two simulated plumes, we also have two real plumes of SF$_6$ and NH$_3$.
All scenes are LWIR captures of the Los Angeles basin area with 128 sampled wavelengths from $7.56-13.16\mu\text{m}$.
Figure \ref{fig:plumes} shows the four plume ROIs with context from the false colored image.

There are several hyperparameters that will impact the performance of the local whitening process.
They are the TED$_\gamma$ (range 0-100\%), TAL$_\beta$ (range 0-100\%), iBATE, and $\min K$ (from 1 to 1024) which is the minimum number of additional non-plume pixels to include during the iBATE process.
Note that if iBATE is off then the initial local background estimate is used, i.e. the average of the most similar non-plume pixels found using TAL and TED.
It would be ideal if the optimal values for all these hyperparameters were roughly equal across the different plumes, as this would mean hyperparameters do not need to be tuned on a per plume basis.
We find the optimal values for each plume by conducting hyperparameter searches.
For the simulated plumes, we search for parameters that minimize the MSE between the locally whitened plume and the whitened ground truth signal.
For the real plumes, we find parameters that maximize the prediction confidence of our deep learning identification model for the correct gas.

\begin{figure*}[htb]
    \centering
    \includegraphics[width=.95\textwidth]{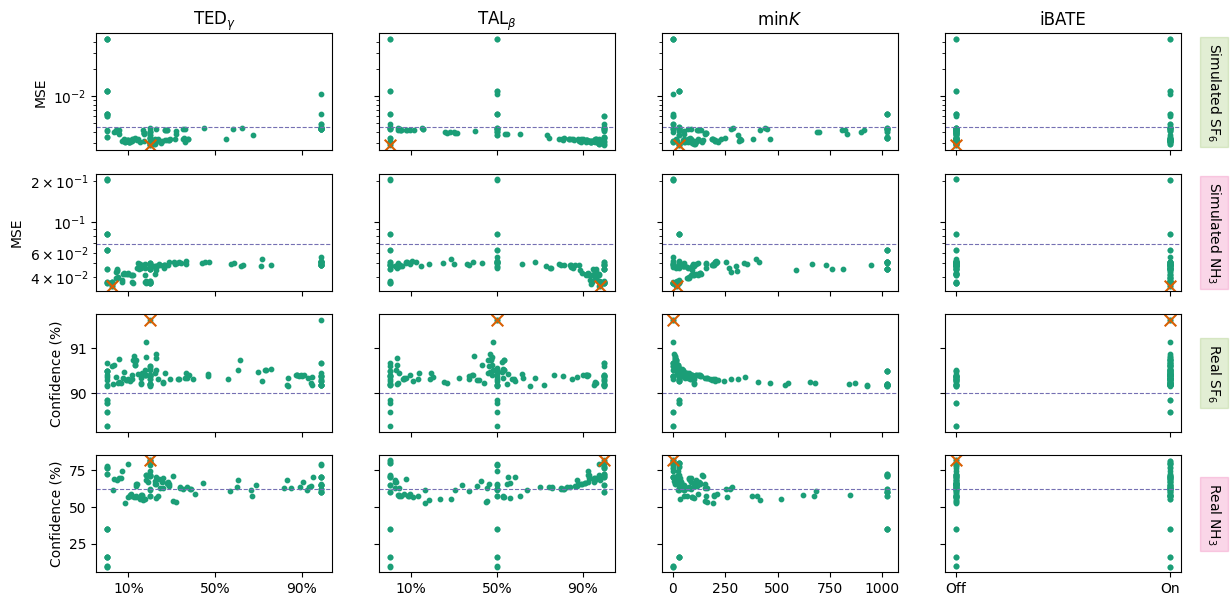}
    \caption{Hyperparameter search slice plots. The orange x marks the optimal hyperparameters, and the horizontal line marks the baseline performance from global whitening.
    The first two rows are the simulated gases where lower MSE is preferred. The last two rows are the real plumes, where higher confidence is preferred.}
    \label{fig:hps}
\end{figure*}

We use \texttt{Optuna} to conduct a grid search to explore the range of the hyperparameter space (54 trials), followed by a Tree-structured Parzen Estimator (TPE) search to narrow down to the optimal hyperparameters (74 trials) \cite{Akiba2019OptunaAN}.
Figure \ref{fig:hps} shows the results of the hyperparameter searches, and the best results are summarized in Table \ref{tab:hps}.
There are several conclusions we can make from these figures.
Looking at the first column, we see that the optimal TED$_\gamma$ is about 20\% for each plume.
The general trend shows that a smaller TED$_\gamma$ is optimal, while using TED$_\gamma=0$, which is equivalent to Euclidean distance, is least optimal.

\begin{table}[htb]
    \centering
    \begin{tabular}{r|l|l|l|l}
         & \multicolumn{2}{|c|}{Simulated (MSE)} & \multicolumn{2}{c}{Real (Conf \%)}\\
        \cline{2-5}
        Method  & $\text{SF}_6$ & $\text{NH}_3$ & $\text{SF}_6$ & $\text{NH}_3$\\
        \hline
        Global  & 0.00454 & 0.06932 & 90.00\% & 62.06\%\\
        LEBEAUS & 0.00282 & 0.03423 & 91.62\% & 81.87\%
    \end{tabular}
    \caption{Best performance for each plume and method. Lower MSE is preferred for the simulated plumes, and higher confidence is preferred for the real plumes.}
    \label{tab:hps}
\end{table}

The second column shows the optimal TAL$_\beta$.
For the simulated plumes, the clear trend is that a high TAL$_\beta$ is better than a small TAL$_\beta$.
In other words, Single Linkage is better than Average Linkage.
Though the real SF$_6$ plume prefers TAL$_\beta=50\%$, generally a larger $\beta$ is preferable to a smaller $\beta$.
The most important takeaway is that the optimal values are similar across all four plumes.

The third column shows optimal values for $\min K$, and there is a clear trend: including fewer pixels consistently outperforms including many additional pixels for iBATE.
This suggests that if too many additional segments are considered, the diversity of the segments increases leading to a worse estimate of the specific local background.

Lastly, the fourth column shows that iBATE typically does not perform significantly worse than not using iBATE, and that in some situations it offers large improvements (e.g. row 3 of Figure \ref{fig:hps}).
From this small sampling of plumes, most hyperparameters outperform global whitening and there are similar optimal hyperparameters across the different plumes.
This suggests that for a variety of plumes, little tuning will be required to obtain satisfactory performance.

\section{Conclusions}
\label{sec:conclusions}

Machine learning for automatic plume detection and identification is becoming more important as the number of images, the resolution of images, and the library of candidate gases increase.
Since real world plumes in urban scenes typically cover several background materials, it is invaluable to appropriately characterize each background material.
With appropriate background estimates, gas spectra can be better estimated, leading to increased confidence when models or spectral analysts identify the gas.

Our LEBEAUS method is able to locally estimate each different background material under a plume.
Using image segmentation to break the image into different background materials, identifying sets of similar segments, then iteratively estimating the local background signatures, we demonstrate LEBEAUS's superiority compared to standard global whitening on both simulated and real plumes.
Furthermore, we find that the optimal hyperparameters are similar across both simulated and real plumes, and among plumes of different gases.
This suggests that a set of ``universal" default hyperparameters can be applied to any plume and it will likely outperform global whitening.

Several important studies are needed to understand the robustness and generalizability of these results.
First, robust and optimal distance functions for clustering segments, as well as alternative linkage functions and segmentation methods, exist that may further improve iterative background estimates.
Second, expanding the study to a greater variety of gas species and plume morphologies is needed to determine how well the iterative method converges with variation in parameters such as plume size, gas concentration, and background heterogeneity.

\bibliographystyle{IEEEbib}
\bibliography{refs}

\end{document}